\documentclass{article}

\usepackage{PRIMEarxiv}

\usepackage[utf8]{inputenc} % allow utf-8 input
\usepackage[T1]{fontenc}    % use 8-bit T1 fonts
\usepackage{url}            % simple URL typesetting
\usepackage{booktabs}       % professional-quality tables
\usepackage{amsfonts}       % blackboard math symbols
\usepackage{nicefrac}       % compact symbols for 1/2, etc.
\usepackage{microtype}      % microtypography
\usepackage{lipsum}
\usepackage{color}
\usepackage{subcaption}
\usepackage[dvipsnames]{xcolor}
\usepackage{pdfpages}
\usepackage{fancyhdr}       % header
\usepackage{graphicx}       % graphics
\usepackage{amsmath,amssymb,amsfonts}
\usepackage{algorithm} 
\usepackage{algorithmic}  
\usepackage[algo2e]{algorithm2e} 
\usepackage{tabularx} % Include this in your document preamble
\graphicspath{{media/}}     % organize your images and other figures under media/ folder
\usepackage[colorlinks,linkcolor=pink]{hyperref}

\title{Enhanced Safety in Autonomous Driving: Integrating Latent State Diffusion Model for End-to-End Navigation
}

\author{
  Detian Chu$^{\ast}$\\
  detian\_chu@aeu.edu.cn
   \And
  Linyuan Bai$^{\ast}$ \\ 
  linyuan\_bai@aeu.edu.cn
   \And
  Jianuo Huang \\
  arnohuangjianuo@gmail.com
   \And
  Zhenlong Fang \\
  1721710605@qq.com
   \And
  Peng Zhang \\
  shuiyuantou2519@163.com
   \And
  Wei Kang \\
  w\_kang2023@163.com
   \And
  Haifeng Lin$^{\dagger}$ \\
  hfling@compintell.cn
}

\begin{document}
\maketitle 

\renewcommand{\thefootnote}{}
\footnotetext{$^{\dagger}$ Corresponding author: Haifeng Lin, hfling@compintell.cn}
\footnotetext{$^{\ast}$Both authors contributed equally to this research.}
\begin{abstract}
With the advancement of autonomous driving, ensuring safety during motion planning and navigation is becoming more and more important. However, most end-to-end planning methods suffer from a lack of safety. This research addresses the safety issue in the control optimization problem of autonomous driving, formulated as Constrained Markov Decision Processes (CMDPs). We propose a novel, model-based approach for policy optimization, utilizing a conditional Value-at-Risk based Soft Actor Critic to manage constraints in complex, high-dimensional state spaces effectively. Our method introduces a worst-case actor to guide safe exploration, ensuring rigorous adherence to safety requirements even in unpredictable scenarios. The policy optimization employs the Augmented Lagrangian method and leverages latent diffusion models to predict and simulate future trajectories. This dual approach not only aids in navigating environments safely but also refines the policy's performance by integrating distribution modeling to account for environmental uncertainties. Empirical evaluations conducted in both simulated and real environment demonstrate that our approach outperforms existing methods in terms of safety, efficiency, and decision-making capabilities.
\end{abstract}

% keywords can be removed

\section{Introduction}
In the rapidly evolving field of autonomous driving, ensuring the safety of vehicles during the exploration phase is very important~\cite{yurtsever2020survey}. Traditional end-to-end methods often struggle to guarantee safety, particularly in complex, high-dimensional environments~\cite{kuutti2022end,xiao2020multimodal}. The increased complexity of the state space in such scenarios not only makes the sampling and learning processes inefficient but also complicates the pursuit of globally optimal policies. The outcomes of inadequate safety measures can be severe, ranging from irreversible system damage to significant threats to human life.

Reinforcement Learning (RL)  has achieved significant success across various fields~\cite{kiran2021deep,mnih2015human}. Furthermore, Deep  learning (DL) is known for its strong perception ability, while reinforcement learning (RL) excels in decision-making. By combining the strengths of DL and RL, deep reinforcement learning (DRL) offers a solution to the decision-planning problem in complex obstacle avoidance scenarios. Unlike traditional motion planning methods, DRL can enhance the adaptability and generalization ability across different scenarios, thus overcoming the limitations of traditional methods and providing a more efficient and effective solution.
Mnih et al.proposed a deep Q-network model that combines convolutional neural networks and Q-learning from traditional RL to address high-dimensional perception-based decision problems~\cite{mnih2015human}. They employed a deep Q-network (DQN) to evaluate the Q-function for Q-learning. This approach has been widely adopted and serves as the primary driver for deep RL.
DRL  empowers robots with perceptual and decision-making abilities by processing input data to yield the output in an end-to-end manner \cite{kim2018review}. The end-to-end motion planning method treats the system as a whole, making it more robust \cite{achiam2017constrained,alan2023control}. Furthermore, deep reinforcement learning can handle high-dimensional and nonlinear environments by leveraging neural networks to learn complex state and action spaces\cite{verstraete2019deep,hao2023latent}. In contrast, traditional heuristic algorithms require manual design of the state and action spaces and may need rules to be redesigned when encountering new scenarios, leading to algorithm limitations and performance bottlenecks.  Previous work in reinforcement learning (RL) and deep learning (DL) has demonstrated advancements in handling complex scenarios, but traditional methods face limitations due to their reliance on manual design for state and action spaces, which cannot adapt well to new scenarios and often result in performance bottlenecks and adaptability issues.

% \begin{figure}
%   \centering
%   \includegraphics[width=12cm]{structure.png}
%   \caption{An overview of our proposed framework. Enhanced safe navigation  Autonomous Driving model with Latent State End-to-Navigation Diffusion model (ESAD-LEND) The process initiates as the agent takes an action in the environment and subsequently receives the state $S$ and reward $r$. The state is encoded into a latent space $Z$ using an encoder. This latent representation is then decoded to simulate environmental states for decision-making. The policy $\pi(a|s, Z)$ leverages these latent representations to determine appropriate actions $a_1, a_2, \ldots, a_n$ across states $S_1, S_2, \ldots, S_n$.}
%   \label{fig:fig1}
% \end{figure}

Recognizing these challenges, several approaches have been proposed. One of them is  the framework of constrained Markov Decision Processes (CMDPs), which have aimed to strike a balance between reward maximization and risk minimization by optimizing the trade-off between exploration and exploitation~\cite{wei2011point,shalev2016safe}. Building on these previous works, this study redefines the control optimization problem within the context of CMDPs. 
We introduce a novel, model-based policy optimization approach that incorporates the Augmented Lagrangian method, and latent diffusion models, specifically designed to address safety constraints effectively in autonomous navigation tasks, meanwhile, optimizing the navigation ability.

Our methodology initiates with a latent variable model aimed at producing extended-horizon trajectories, thereby boosting our system’s ability to predict future states more accurately. To further strengthen the safety of our model, we incorporate a conditional Value-at-Risk (VaR) within the Soft Actor-Critic (SAC) framework, which ensures that safety constraints are met through the use of the Augmented Lagrangian method for efficiently solving the safety-constrained optimization challenges.

Furthermore, in tackling scenarios of extreme risk, our approach includes a worst-case scenario planning method that leverages a "worst-case actor" during policy exploration to ensure the safest possible outcomes even in potentially dangerous conditions. To enhance our model's performance, we have integrated a latent diffusion model for state representation learning, refining our system's ability to interpret complex environmental data.

\begin{figure}[htbp]
\centering
\includegraphics[width=0.7\linewidth]{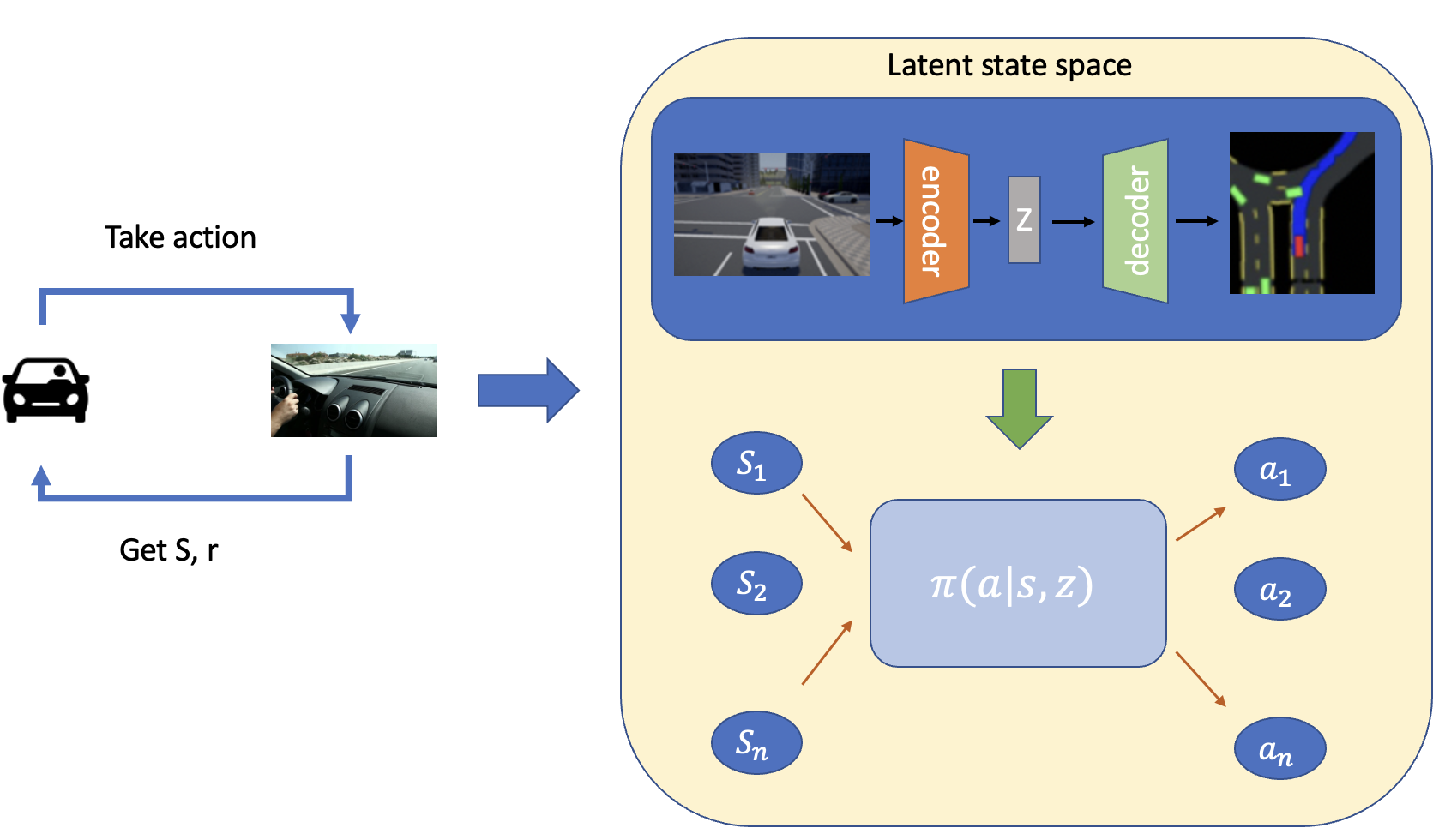}
\caption{An overview of our proposed framework. Enhanced safe navigation  Autonomous Driving model with Latent State End-to-Navigation Diffusion model (ESAD-LEND) The process initiates as the agent takes an action in the environment and subsequently receives the state $S$ and reward $r$. The state is encoded into a latent space $Z$ using an encoder. This latent representation is then decoded to simulate environmental states for decision-making. The policy $\pi(a|s, Z)$ leverages these latent representations to determine appropriate actions $a_1, a_2, \ldots, a_n$ across states $S_1, S_2, \ldots, S_n$.}
\label{fig:fig1}
\end{figure}

The efficacy of our proposed approach is validated through extensive experiments. Our empirical findings demonstrate the capability of our approach to manage and mitigate risks effectively in high-dimensional state spaces, thereby enhancing the safety and reliability of autonomous vehicles in complex driving environments.

Our main contributions are listed below:

\begin{itemize}
    \item We introduce latent diffusion models for state representation learning, which enable the forecasting of future observations, rewards, and actions. This capability allows for the simulation of future trajectories directly within the model framework, facilitating proactive assessment of rewards and risks through controlled model roll-outs. 
    
    \item We extend our approach to include advanced prediction of the value distributions of future states, incorporating the estimation of worst-case scenarios which ensures that our model not only predicts but also prepares for potential adverse conditions, enhancing system reliability.
    
    % \item We guarantee that all modeled scenarios remain within established safety parameters. This enhancement improves the reliability of the autonomous system under challenging conditions, safeguarding the system against unexpected events and maintaining safety as a paramount concern during operational deployment.
    
    \item Our experimental results demonstrate the effectiveness of our proposed methods through testing in both simulated environments and real-world settings. 
\end{itemize}

\section{Related work}

\subsection{Safe Reinforcement Learning}
Safe Reinforcement Learning (Safe RL) is a method that combines safety measures with the usual learning process for agents in complex environments ~\cite{garcia2015comprehensive}. The development of safe and reliable autonomous driving systems requires a comprehensive focus on safety throughout the control and decision-making processes. In this context, safe reinforcement learning (Safe RL) emerges as a powerful tool for training controllers and planners capable of navigating dynamic environments while adhering to safety constraints.

Safe RL algorithms address the safety challenge through various approaches. Notable examples include Constrained Policy Optimization (CPO)~\cite{achiam2017constrained}, which optimizes policies within explicitly defined safety constraints to prevent unsafe actions. Trust Region Policy Optimization (TRPO)~\cite{schulman2015trust} and Proximal Policy Optimization (PPO)~\cite{schulman2017proximal} adapt these algorithms to incorporate safety considerations by introducing penalty terms or barrier functions. Gaussian Processes (GPs) model~\cite{deisenroth2013gaussian} environmental uncertainty, enabling safe exploration by quantifying the risk associated with different actions. Model Predictive Control (MPC)-based Safe RL~\cite{koller2018learning}~\cite{zanon2020safe} predicts future states and optimizes actions over a finite horizon while adhering to safety constraints.

In the context of autonomous driving, Safe RL plays a crucial role in ensuring safety. Applications include high-level decision-making for tasks like lane changing~\cite{shalev2016safe} and route planning~\cite{ye2020automated}, considering the behavior of surrounding vehicles and potential conflicts. Low-level control integrates Safe RL with traditional control methods like MPC, ensuring adherence to safety constraints while following the planned path~\cite{hewing2020learning}~\cite{brudigam2021stochastic}. The combination of RL with rule-based systems encoding traffic laws guarantees that learned policies comply with regulations. Safety verification techniques provide guarantees that learned policies remain safe even in the presence of uncertainties and dynamic obstacles. ~\cite{pek2017verifying}~\cite{wang2019lane}

The diverse Safe RL algorithms and their applications underscore their critical role in developing adaptable and safe control systems. Continuous advancements in these methods will be crucial for the successful real-world deployment of autonomous vehicles.

\subsection{Reinforcement Learning with Latent State}
\label{preliminary}
Latent dynamic models, which represent a cornerstone in the modeling of time-series data within the field of reinforcement learning (RL), facilitate a profound understanding of hidden states and dynamic changes in complex environments~\cite{kim2018review, sewell2015latent, sarkar2005dynamic}. These models capture essential relationships between unobservable internal states and observed data, significantly enhancing the predictive capabilities of RL agents.

In the RL context, latent dynamic models operate under probabilistic frameworks, employing Bayesian inference or maximum likelihood estimation to deduce both model parameters and hidden states with increased accuracy~\cite{padakandla2021survey, levine2018reinforcement}. These frameworks enable the modeling of environments in a way that aligns with the probabilistic nature of real-world dynamics.

Mathematically, latent dynamic models are characterized by:
\begin{itemize}
    \item \textbf{State Transition Equation:} 
    \[
    s_{t+1} = f(s_t, a_t, \epsilon_t)
    \]
    where \( s_t \) denotes the latent state at time \( t \), \( a_t \) the action taken, and \( \epsilon_t \) the stochasticity inherent in the environment. The function \( f \) may be deterministic or stochastic, encapsulating the uncertainty of state transitions.
    
    \item \textbf{Observation Equation:} 
    \[
    o_t = g(s_t, \delta_t)
    \]
    where \( o_t \) represents the observed output linked to the hidden state \( s_t \), with \( \delta_t \) encapsulating the observation noise, thus linking theoretical models to real-world observations.
    
    \item \textbf{Reward Function:} 
    \[
    r_t = R(s_t, a_t)
    \]
    defining the immediate reward received after executing action \( a_t \) in state \( s_t \), essential for policy optimization in RL.
\end{itemize}

The primary goal of employing latent dynamic models in RL is to infer the sequence of hidden states \( s_1, s_2, ..., s_T \) based on observations \( o_1, o_2, ..., o_T \), actions \( a_1, a_2, ..., a_T \), and corresponding rewards \( r_1, r_2, ..., r_T \). This comprehensive modeling approach not only predicts future states or actions but also integrates environmental dynamics to enhance decision-making processes and optimize policy outcomes~\cite{lee2020context, chen2021interpretable, hao2023latent}. Furthermore, these models facilitate a deeper understanding of environmental intricacies, thereby empowering RL agents to make more informed and effective decisions~\cite{li2017inferring, wang2021interpretable}.

\subsection{Diffusion model-based RL}

Recent advances in diffusion models have found significant applications in reinforcement learning (RL), particularly in the domain of offline RL where interaction with the environment is constrained. Notable for their success in generative tasks, diffusion models are adapted to address critical challenges in RL such as the distributional shift and extrapolation errors commonly observed in offline settings \cite{gulcehre2020addressing,ma2021conservative}. 

Studies like \cite{zhang2024predicting} demonstrate that diffusion probabilistic models can effectively generate plausible future states and actions, facilitating robust policy learning from static datasets. Furthermore, latent diffusion models (LDMs) reduce computational demands and enhance learning efficiency by encoding trajectories into a compact latent space before diffusion, thus capturing complex decision dynamics \cite{rombach2022high}. These methods have improved the stability and efficacy of Q-learning algorithms by ensuring that generated actions remain within the behavioral policy's support, thus mitigating the risk of policy deviation due to poor sampling \cite{kumar2020conservative}. 

This integration of diffusion techniques into RL frameworks represents a promising frontier for developing more capable and reliable autonomous systems, particularly in environments where traditional learning methods fall short.

\section{Problem Modeling}
In the problem of autonomous navigation, we model the interaction of an autonomous agent with a dynamic and uncertain environment using a finite-horizon Markov Decision Process (MDP), defined by the tuple $\mathcal{M} \sim (\mathcal{S}, \mathcal{O}, \mathcal{A}, \mathcal{P}, r, \gamma)$. Here, $\mathcal{S} \subset \mathbb{R}^n$ represents a high-dimensional continuous state space, and $\mathcal{A} \subset \mathbb{R}^m$ denotes the action space. State transitions are defined by $s_{t+1} \sim \mathcal{P}(\cdot \mid s_t, a_t)$, reflecting the stochastic nature of the environment.

Observations $\mathcal{O}$, derived from the state space, are typically high-dimensional images captured by sensors and processed via a latent diffusion model to enhance state representation. This allows for a more comprehensive understanding of environmental dynamics, which is important for navigating complex scenarios. The reward function $r: \mathcal{S} \times \mathcal{A} \times \mathcal{S} \rightarrow \mathbb{R}$ and the discount factor $\gamma \in[0,1]$ guide the agent's policy $\pi_\theta$, which generates actions based on processed observations $o_t$.

We emphasize safety through the integration of a conditional Value-at-Risk (VaR) metric within our decision-making framework, addressing the worst-case scenarios with a novel approach that involves latent state diffusion for robust state representation learning. Safety mechanisms are formalized through a subset $S_u \subset \mathbb{R}^n$, where entering a state $s_t \in S_u$ indicates a potential safety violation, monitored by a safety function $\kappa$. The objective extends beyond maximizing cumulative rewards to ensuring minimal safety violations:

\begin{equation}
\max _\theta J\left(\pi_\theta\right) = \mathbb{E}_{\mathcal{P}\left(\cdot \mid s_t, a_t\right)}\left[\sum_{t=0}^T \gamma^t r\left(s_t, a_t, s_{t+1}\right)\right], \text{ s.t. }, \sum_{t=0}^T \kappa\left(s_t\right) \leq D, a_t \sim \pi_\theta\left(\cdot \mid o_t\right).
\end{equation}

Here, $\kappa(s_t) \in \{0,1\}$ serves as an indicator of safety violations, with $D \in \mathbb{R}$ representing the maximum allowable safety violations, aiming for $D \rightarrow 0$ to enhance operational safety. This safety-constrained MDP framework allows the agent to learn navigation policies that maximize efficiency and ensure safety, effectively balancing between achieving high performance and adhering to critical safety constraints.

\section{Methodology}

\subsection{Constrained Markov Decision Process formulation}
Reinforcement Learning is the process of continuous interaction between the agent and the environment. In this study, we focus on the safety of autonomous driving, which requires a trade-off between reward and safety. We formulate the control optimization problem as a Constraint Markov Decision Process (CMDP). CMDP is described by $\left(S, A, p, r, c, d, \gamma\right)$, including the state space, action space, transition model, reward, cost, constraint, and discount factor. Each step, the agent receives reward $r$ and cost $c$, the optimization target is to maximize the reward in the safety constraint in Equation \ref{eq:CMDP}.

\begin{equation}
    \begin{aligned}
    & \max_\pi \underset{\left(s_t, a_t\right) \sim \tau_\pi}{\mathrm{E}}\left[\sum_t \gamma^t r\left(s_t, a_t\right)\right] \\
    & \text{s.t.} \underset{\left(s_t, a_t\right) \sim \tau_\pi}{\mathrm{E}}\left[\sum_t \gamma^t c\left(s_t, a_t\right)\right] \leq d
    \end{aligned}
\label{eq:CMDP}
\end{equation}

where $\pi$, $\tau_{\pi}$ denotes for the policy and the trajectory distribution of policy.

\subsection{Build the Latent Diffusion Model for State Representation}

% \begin{figure}
%   \centering
%   \includegraphics[width=12cm]{train-diffusion.png}
%   \caption{Illustration of the training process for a diffusion-based model. Starting with the latent state \( Z_0 \), the model undergoes a forward diffusion process, progressively adding noise through multiple steps until it reaches the fully noised state \( Z_T \). This process is reversed by a denoising network, which iteratively refines \( Z_T \) back to \( Z_0 \) by gradually removing noise across several iterations, labeled here as \( Z_{j-1} \) to \( Z_0 \). The cycle between the forward diffusion and denoising processes is repeated multiple times to enhance the model's ability to accurately regenerate the original latent state from its noised version.}
%   \label{fig:fig2}
% \end{figure}

\begin{figure}[htbp]
\centering
\includegraphics[width=0.7\linewidth]{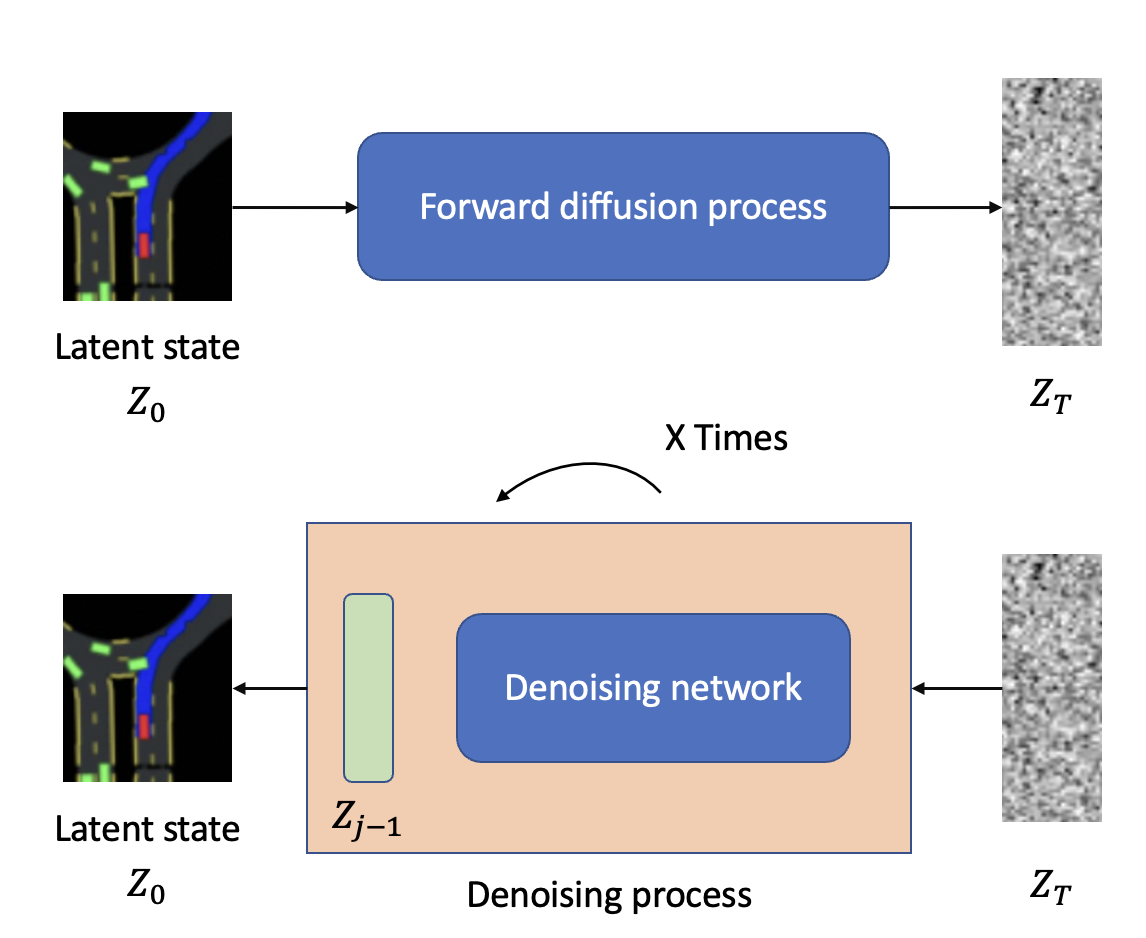}
\caption{The policy generation process in a diffusion-based control system. Starting with a state representation of the environment, the diffusion model \( P(z|s) \) generates multiple candidate latent states. These candidates are evaluated by a Q-function \( \arg\max Q(s, z) \), selecting the optimal latent state for action execution. The process integrates a safety guarantee mechanism to ensure that the chosen action meets predefined safety criteria before execution, culminating in a controlled action that adheres to safety standards.}
\label{fig:fig2}
\end{figure}

% \begin{figure}[H]
% \begin{adjustwidth}{-\extralength}{0cm}
% \centering
% \includegraphics[width=10cm]{train-diffusion.png}
% \end{adjustwidth}
% \caption{Illustration of the training process for a diffusion-based model. Starting with the latent state \( Z_0 \), the model undergoes a forward diffusion process, progressively adding noise through multiple steps until it reaches the fully noised state \( Z_T \). This process is reversed by a denoising network, which iteratively refines \( Z_T \) back to \( Z_0 \) by gradually removing noise across several iterations, labeled here as \( Z_{j-1} \) to \( Z_0 \). The cycle between the forward diffusion and denoising processes is repeated multiple times to enhance the model's ability to accurately regenerate the original latent state from its noised version.}
% \end{figure}  

Our model designs a sophisticated latent model consisting of three primary components: a representation model, a transition model, and a reward model. Each component plays a crucial role in the system's ability to predict and navigate complex environments by learning from both observed and imagined trajectories.

\begin{itemize}
  \item \textbf{Representation Model:} The representation model is crucial for establishing a robust latent space based on past experiences. It is formalized as $p(s_\tau \mid s_{\tau-1}, a_{\tau-1}, o_\tau)$, where the model predicts the next state by synthesizing information from the current state, action, and observation. The loss in representation is quantified by assessing the accuracy of state and reward predictions.

  \item \textbf{Transition Model:} This model outputs a Gaussian distribution from which the next state is sampled, defined as $q(s_\tau \mid s_{\tau-1}, a_{\tau-1})$. The transition model's accuracy is evaluated using the Kullback-Leibler (KL) divergence between the prior and posterior distributions, representing the latent imagination and the environment's real response, respectively.

  \item \textbf{Reward Model:} The reward model optimizes the learning process by calculating expected rewards based on the current state, expressed as $q(r_\tau \mid s_\tau)$. This model is pivotal for the agent to learn and optimize actions that maximize returns from the environment.
\end{itemize}

In our framework, $p$ denotes the distribution sampling from environment interactions, while $q$ represents predictions generated from the latent imagination. We use $\tau$ to index time steps within the latent space. The core innovation of our model lies in its ability to create and refine a latent imagination space that predicts future trajectories. This capability enables the agent to safely explore and learn optimal behaviors by iteratively refining the latent space to avoid unsafe states and maximize efficiency.

\subsection{Build Safety Guarantee}

% \begin{figure}
%   \centering
%   \includegraphics[width=3cm]{policy.png}
%   \caption{The policy generation process in a diffusion-based control system. Starting with a state representation of the environment, the diffusion model \( P(z|s) \) generates multiple candidate latent states. These candidates are evaluated by a Q-function \( \arg\max Q(s, z) \), selecting the optimal latent state for action execution. The process integrates a safety guarantee mechanism to ensure that the chosen action meets predefined safety criteria before execution, culminating in a controlled action that adheres to safety standards.}
%   \label{fig:fig3}
% \end{figure}

\begin{figure}[htbp]
\centering
\includegraphics[width=0.4\linewidth]{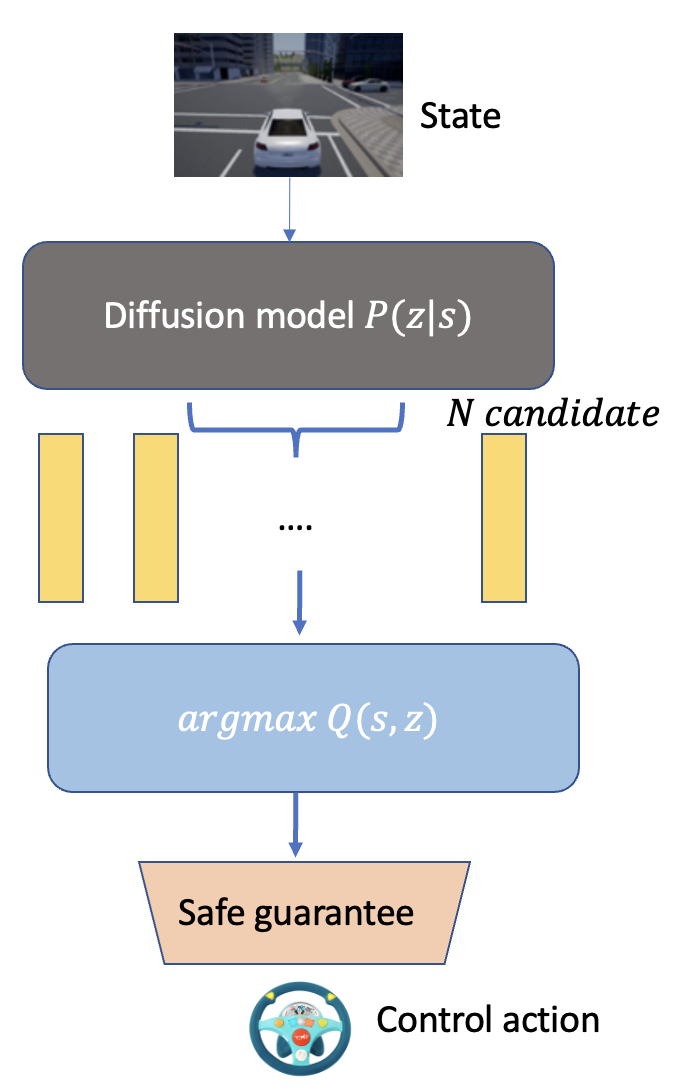}
\caption{The policy generation process in a diffusion-based control system. Starting with a state representation of the environment, the diffusion model \( P(z|s) \) generates multiple candidate latent states. These candidates are evaluated by a Q-function \( \arg\max Q(s, z) \), selecting the optimal latent state for action execution. The process integrates a safety guarantee mechanism to ensure that the chosen action meets predefined safety criteria before execution, culminating in a controlled action that adheres to safety standards.}
\label{fig:fig3}
\end{figure}

Figure \ref{fig:fig3} depicts the policy generation process within a diffusion-based control system. The process begins with the state of the environment, which is then input into a diffusion model denoted as $P(z|s)$. This model generates multiple candidate latent states, which represent potential actions or decision paths the system might take.

These candidates are then evaluated using a Q-function, $\arg\max Q(s, z)$, which selects the optimal latent state for action execution based on maximizing expected rewards and ensuring compliance with safety standards. The chosen action is not only optimal in terms of performance but also satisfies predefined safety criteria, ensuring that the control action adheres to safety standards before execution.

This structured approach ensures a robust safety guarantee mechanism, integrating both performance optimization and rigorous safety compliance, which is crucial for autonomous systems operating in dynamic and uncertain environments.

Similar approach can be found in Latent imagination e.g., Dreamer ~\cite{hafner2019dream}, and got great performance. However, Dreamer did not take safety constraints into consideration. Though Dreamer achieved high reward target, without safety, the agent might cause irreversible incident in order to reach higher reward. It is significant to introduce the constraints of safety to balance reward and risk.

In the latent imagination, we leverage the distributional reinforcement learning to solve the safety-constrained RL problem ~\cite{yang2021wcsac}. Instead of following a policy $\pi$ to get an action value, distributional RL considers on the expectation of value and cost ~\cite{mavrin2019distributional}. We focus on the actor-critic policy and build a model-based algorithm together with the latent imagination. Soft actor-critic (SAC) ~\cite{haarnoja2018soft} introduces the concept of maximum entropy balancing the exploitation and exploration:  

\begin{equation}
\pi^*=\underset{\pi}{\operatorname{argmax}} \sum_{t=0}^T E_{\left(s_t, a_t\right) \sim \tau_\pi}\left[\gamma^t\left(r\left(s_t, a_t\right)+\beta \mathcal{H}\left(\pi\left(. \mid s_t\right)\right)\right]\right.
\end{equation}

where  $\pi^*$ denotes the optimal policy, $\beta$ denotes the stochasticity of $\pi$.

Agents can learn the optimal policy without safety concern, however, irreversible situation such as collision cannot be accepted in autonomous driving. We consider the safety constraints and formulate it as a Lagrangian method with constraints:

\begin{equation}
\begin{aligned}
& \max_{\pi} \underset{\left(s_t, a_t\right) \sim \rho_\pi}{\mathbb{E}}\left[\sum_t \gamma^t r\left(s_t, a_t\right)\right] \\
\text{s.t.} & \left\{
\begin{array}{l}
\underset{\left(s_t, a_t\right) \sim \rho_\pi}{\mathbb{E}}\left[\sum_t \gamma^t c\left(s_t, a_t\right)\right] \leq d \\
\underset{\left(s_t, a_t\right) \sim \rho_\pi}{\mathbb{E}}\left[-\log \left(\pi_t\left(a_t \mid s_t\right)\right)\right] \geq \mathcal{H}_0 \quad \forall t \\
h(s_t) \leq 0 \quad \text{(Safety Constraint)} \\
h(s_t+1) \leq (1 - \alpha)h(s_t) \quad \text{(Control Barrier Function)}
\end{array}
\right.
\end{aligned}
\end{equation}

To better use the constraints of safety, barrier function is also leveraged to modify the risk value in distributional RL so that the agent can evaluate when to explore more and when to be conservative.  In this formulation, For a constraint with relative degree $\mathrm{m}$, the generalized control barrier function is:
$$
h\left(s_{t+m}\right) \leq(1-\alpha)h\left(s_t\right)
$$
For a certain risk level $\alpha$, we optimize policy until $\Gamma_\pi$ satisfies:
$$
\begin{gathered}
\Gamma_\pi(s, a, \alpha) \doteq C V a R_\alpha=Q_\pi^c(s, a)+\alpha^{-1} \phi\left(\Phi^{-1}(\alpha)\right) \sqrt{V_\pi^c(s, a)} \\
\Gamma_\pi\left(s_{t+m}, a_{t+m}, \alpha\right) \leq(1-\alpha)\Gamma_\pi\left(s_t, a_t, \alpha\right)
\end{gathered}
$$
where $\alpha$ is the conservativeness coefficient.

%\(h(s_t)\) represents the safety constraint, and the control barrier function is included as 
% \[h(s_{t+1}) \leq (1 - \alpha)h(s_t).\]
% The parameter \(\alpha\) controls the conservativeness of the constraints. A larger \(\alpha\) makes the constraints less conservative.

\textbf{Critic}, Instead of $Q_\pi^c(s, a)$ is the expectation of long-term cumulative costs from starting point $(s, a)$, denoted by:
$$
C_\pi(s, a)=\sum_t \gamma_t c\left(s_t, a_t\right) \mid s_0=s, a_0=a, \pi
$$
Following policy $\pi$,the probability distribution of $C_\pi(s, a)$ is modeled as $p^\pi(C \mid s, a)$, such that:
$$
Q_\pi^c(s, a)=\mathbb{E}_{p^\pi}[C \mid s, a].
$$

The distributional Bellman Operator $\mathcal{T}^\pi$ is defined as:
$$
\mathcal{T}^\pi C(s, a)=c(s, a)+\gamma C\left(s^{\prime}, a^{\prime}\right)
$$
$s^{\prime} \sim p(\cdot \mid s, a), a^{\prime} \sim \pi\left(\cdot \mid a^{\prime}\right)$. We approximate the distribution $C_\pi(s, a)$ with a Gaussian distribution $C_\pi(s, a) \sim \mathcal{N}\left(Q_\pi^c(s, a), V_\pi^c(s, a)\right)$, where $V_\pi^c(s, a)=\mathbb{E}_{p^\pi}\left[C^2 \mid s, a\right]-Q_\pi^c(s, a)^2$.

- To estimate $Q_\pi^c$, we can use Bellman function:
$$
Q_\pi^c(s, a)=c(s, a)+\gamma \sum_{s^{\prime} \in S} p\left(s^{\prime} \mid s, a\right) \sum_{a^{\prime} \in A} \pi\left(a^{\prime} \mid s^{\prime}\right) Q_\pi^c\left(s^{\prime}, a^{\prime}\right) .
$$
- The projection equation for estimating $V_\pi^c(s, a)$ is:
$$
\begin{gathered}
V_\pi^c(s, a)=c(s, a)^2-Q_\pi^c(s, a)^2+2 \gamma c(s, a) \sum_{s^{\prime} \in S} p\left(s^{\prime} \mid s, a\right) \sum_{a^{\prime} \in A} \pi\left(a^{\prime} \mid s^{\prime}\right) Q_\pi^c\left(s^{\prime}, a^{\prime}\right)+ \\
+\gamma^2 \sum_{s^{\prime} \in S} p\left(s^{\prime} \mid s, a\right) \sum_{a^{\prime} \in A} \pi\left(a^{\prime} \mid s^{\prime}\right) V_\pi^c\left(s^{\prime}, a^{\prime}\right)+\gamma^2 \sum_{s^{\prime} \in S} p\left(s^{\prime} \mid s, a\right) \sum_{a^{\prime} \in A} \pi\left(a^{\prime} \mid s^{\prime}\right) Q_\pi^c\left(s^{\prime}, a^{\prime}\right)^2 .
\end{gathered}
$$
We use two neural networks parameterized by $\mu$ and $\eta$ respectively to estimate the safety-critic:
$$
Q_\mu^c(s, a) \rightarrow \widehat{Q_\pi^c}(s, a) \text { and } V_\eta^c(s, a) \rightarrow \widehat{V_\pi^c}(s, a)
$$
2-Wasserstein distance: $W_2(u, v)=\left\|Q_1-Q_2\right\|_2^2+\operatorname{trace}\left(V_1+V_2-2\left(V_2^{1 / 2} V_1 V_2^{1 / 2}\right)^{1 / 2}\right)$
The 2-Wasserstein distance can be computed as the Temporal Difference (TD) error based on the projection equations $Q_\pi^c(s, a)$ and $V_\pi^c(s, a)$ to update the safety-critic, i.e., we will minimize the following values:
$$
\begin{gathered}
J_C(\mu)=\underset{\left(s_t, a_t\right) \sim \mathcal{D}}{\mathbb{E}}\left\|\Delta Q\left(s_t, a_t, \mu\right)\right\|_2^2 \\
J_V(\eta)=\underset{\left(s_t, a_t\right) \sim \mathcal{D}}{\mathbb{E}} \operatorname{trace}\left(\Delta V\left(s_t, a_t, \mu\right)\right)
\end{gathered}
$$
Where, $\mu$ and $\eta$ are two parameters that represents the two neural network in safety critic, $J_C(\mu)$ is the loss function of $Q_\mu^c(s, a)$ and $J_V(\eta)$ is the loss function of $V_\eta^c(s, a)$. We can get:
$$
\begin{gathered}
\Delta Q\left(s_t, a_t, \mu\right)=\overline{Q_\mu^c}\left(s_t, a_t\right)-Q_\mu^c(s, a) \\
\Delta V\left(s_t, a_t, \mu\right)=\overline{V_\mu^c}\left(s_t, a_t\right)+V_\mu^c\left(s_t, a_t\right)-2\left(V_\mu^c\left(s_t, a_t\right)^{\frac{1}{2}} \overline{V_\mu^c}\left(s_t, a_t\right) V_\mu^c\left(s_t, a_t\right)^{\frac{1}{2}}\right)^{\frac{1}{2}}
\end{gathered}
$$
where $\overline{Q_\mu^C}\left(s_t, a_t\right)$ and $\overline{V_\mu^C}\left(s_t, a_t\right)$ is the TD target.

\textbf{Actor}, We focus on the expectation of cumulative reward and cost to avoid the impact of random sampling from the Gaussian distribution. A parameter $\alpha$ is designed to set the risk attention, denotes the risk-aware level on safety. $\alpha \in\left(0,1\right)$, the smaller scalar means more pessimistic in safety, whereas the larger one expect a less risk-aware state.

We set a $\alpha$-percentile $F_C^{-1}(1-\alpha)$, based on the Conditional Value at Risk (CVaR), we can get $C V a R_\alpha \doteq \mathbb{E}_{p^\pi}\left[C \mid C \geq F_C^{-1}(1-\alpha)\right]$. Given the risk-aware level manually by different situation or standard, the control optimization problem can be modified into: \begin{equation}
\mathbb{E}_p \pi\left[C\left(s_t, a_t\right) \mid C\left(s_t, a_t\right) \geq F_C^{-1}(1-\alpha)\right] \leq d \quad \forall t
\end{equation}

Considering the critic mentioned before, the safety measurements can be changed into:
\begin{equation}
\Gamma_\pi(s, a, \alpha) \doteq C V a R_\alpha=Q_\pi^c(s, a)+\alpha^{-1} \phi\left(\Phi^{-1}(\alpha)\right) \sqrt{V_\pi^c(s, a)}
\end{equation}

As the safety critic, we also define a reward critic with the neural network parameter $\psi$ and loss function $J_R\left(\psi_i\right)$. Given the risk level, we use $\Gamma_\pi(s, a, \alpha) \leq d \quad \forall t$ to find the optimal policy $\pi$. 

\subsection{Value-at-Risk Based
Soft Actor-Critic for Safe Exploration  }

\begin{figure}[ht]
\centering
\includegraphics[width=0.8\textwidth]{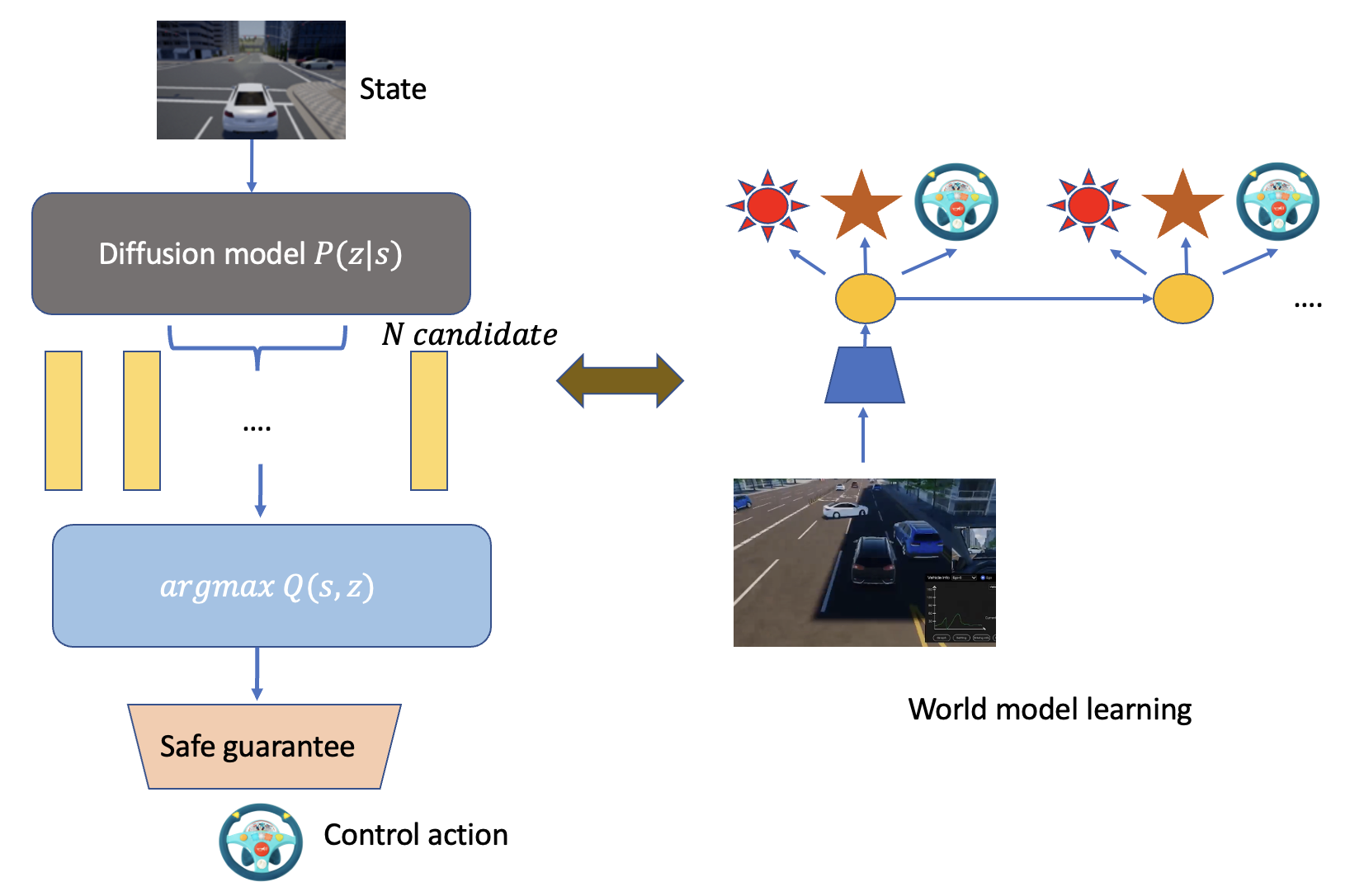}
\caption{Schematic representation of the world model learning process in our autonomous system. The process begins with the state input into the diffusion model $P(z|s)$, which generates multiple candidate latent states. These states are evaluated using a Q-function, $\arg\max Q(s, z)$, to select the optimal state for action execution, ensuring that it meets safety criteria. The selected action is then applied in the world model learning component, which simulates possible future trajectories based on continuous actor-critic policy optimization and updates in response to new data from the environment. This iterative process enhances the agent’s ability to predict and react to dynamic scenarios, integrating safety constraints at every step.}
\label{fig:world_model_learning}
\end{figure}

\begin{algorithm}[htbp]
\caption{Pseudocode for ESAD-LEND}
\begin{algorithmic}[1]
\STATE Inputs: Initial parameters $\alpha, \psi, \mu, \eta, \theta, \tau$
\STATE Initialize: Target networks: $\langle\bar{\psi}, \bar{\mu}, \bar{\eta}\rangle \leftarrow\langle\psi, \mu, eta\rangle$
\STATE Initialize: Dataset $\mathcal{D}$ with $S$ random seed episodes.
\WHILE{not converged \textbf{do}}
\FOR{update step $t=1 . . T$}
    \STATE // World Model learning with Latent Diffusion Representation
    \STATE Sample a sequence batch of $\left\{\left(o_t, a_t, r_t, c_t\right)\right\}_{t=k}^{k+L}$ from $\mathcal{D}.$
    \STATE Encode observed states to latent space: $z_t \sim q_\theta\left(z_t \mid o_t, h_{t-1}\right)$ using a latent diffusion model.
    \STATE Predict next state and reward using latent representations: $h_t=f_\theta\left(h_{t-1}, z_t, a_{t-1}\right)$.
    \STATE Optimize $\theta$ by minimizing the equation $\mathcal{L}(\theta)$.
    \STATE // Behavior learning
    \STATE Imagine trajectories $\left\{\left(s_\tau, a_\tau \right)\right\}_{\tau=t}^{t+H}$ from each $s_t$.
    \STATE Compute safety measure $\Gamma_\pi(s, a, \alpha)$ based on $\Gamma_\pi(s, a, \alpha) \doteq C V a R_\alpha = Q_\pi^c(s, a)+\alpha^{-1} \phi\left(\Phi^{-1}(\alpha)\right) \sqrt{V_\pi^c(s, a)}$
    \STATE $\psi_i \leftarrow \psi_i-\lambda_R \hat{\nabla}_{\psi_i} J_R\left(\psi_i\right) \text { for } i \in\{1,2\}$
    \STATE $\theta \leftarrow \theta-\lambda_\pi \hat{\nabla}_\theta J_\pi(\theta)$
    \STATE $\mu, \eta, \beta, \kappa \leftarrow \mu-\lambda_C \hat{\nabla}_\mu J_C(\mu), \eta-\lambda_V \hat{\nabla}_\eta J_V(\eta), \beta-\lambda_\beta \hat{\nabla}_\beta J_e(\beta), \kappa-\lambda_\kappa \hat{\nabla}_\kappa J_s(\kappa)$
    \STATE // Update target networks weights
    \STATE $\bar{\psi}_i \leftarrow \tau \psi_i+(1-\tau) \bar{\psi}_i \text { for } i \in\{1,2\}$
    \STATE $\bar{\mu},\bar{\eta} \leftarrow \tau \mu+(1-\tau) \bar{\mu}, \tau \eta+(1-\tau) \bar{\eta}$
\ENDFOR
\STATE // Environment interaction
\STATE $o_1 \leftarrow$ env $\cdot$ reset ()
\FOR{time step $t=1 . . T$ do}
    \STATE Update state $s_t$ using latent diffusion model: $s_t \sim p_\theta\left(s_t \mid s_{t-1}, a_{t-1}, z_t\right)$.
    \STATE Compute action $a_t \sim \pi_\theta\left(a_t \mid s_t\right)$ using the action model.
    \STATE Add exploration noise to action.
    \STATE $r_t, o_{t+1} \leftarrow \operatorname{env} \cdot \operatorname{step}\left(a_t\right)$. 
\ENDFOR
\STATE Add experience to dataset $\mathcal{D} \leftarrow \mathcal{D} \cup\left\{\left(o_t, a_t, r_t\right)_{t=1}^T\right\}$.
\ENDWHILE
\end{algorithmic}
\label{new-algo}
\end{algorithm}

As shown in Algorithm 1, we build the latent imagination by updating the representation model, transition model and reward model. We predict the future trajectories based on the transition model, and regard trajectories as latent imagination for continuous actor-critic policy optimization. Given a specific risk-aware level, new safety constraint is set. Considering the expectation of cumulative cost and reward, we can update the model by distributional RL. After continuous learning in the latent imagination for some discrete time step, the agent interacts with environment and gets new reward and cost data.

%%%%%%%%%%%%%%%%%%%%%%%%%%%%%%%%%%%%%%%%%%
\section{Experiments}

\subsection{Environmental Setup}

\subsubsection{Experimental Setup in CARLA Simulator}
\label{subsec:experimental_setup}

In our study, we utilized the CARLA simulator to construct and evaluate various safety-critical scenarios that challenge the response capabilities of autonomous driving systems. CARLA provides a rich, open-source environment tailored for autonomous driving research, offering realistic urban simulations.

\subsubsubsection{\textbf{Specific scenario}}

\begin{figure}[ht]
\centering
\includegraphics[width=0.9\textwidth]{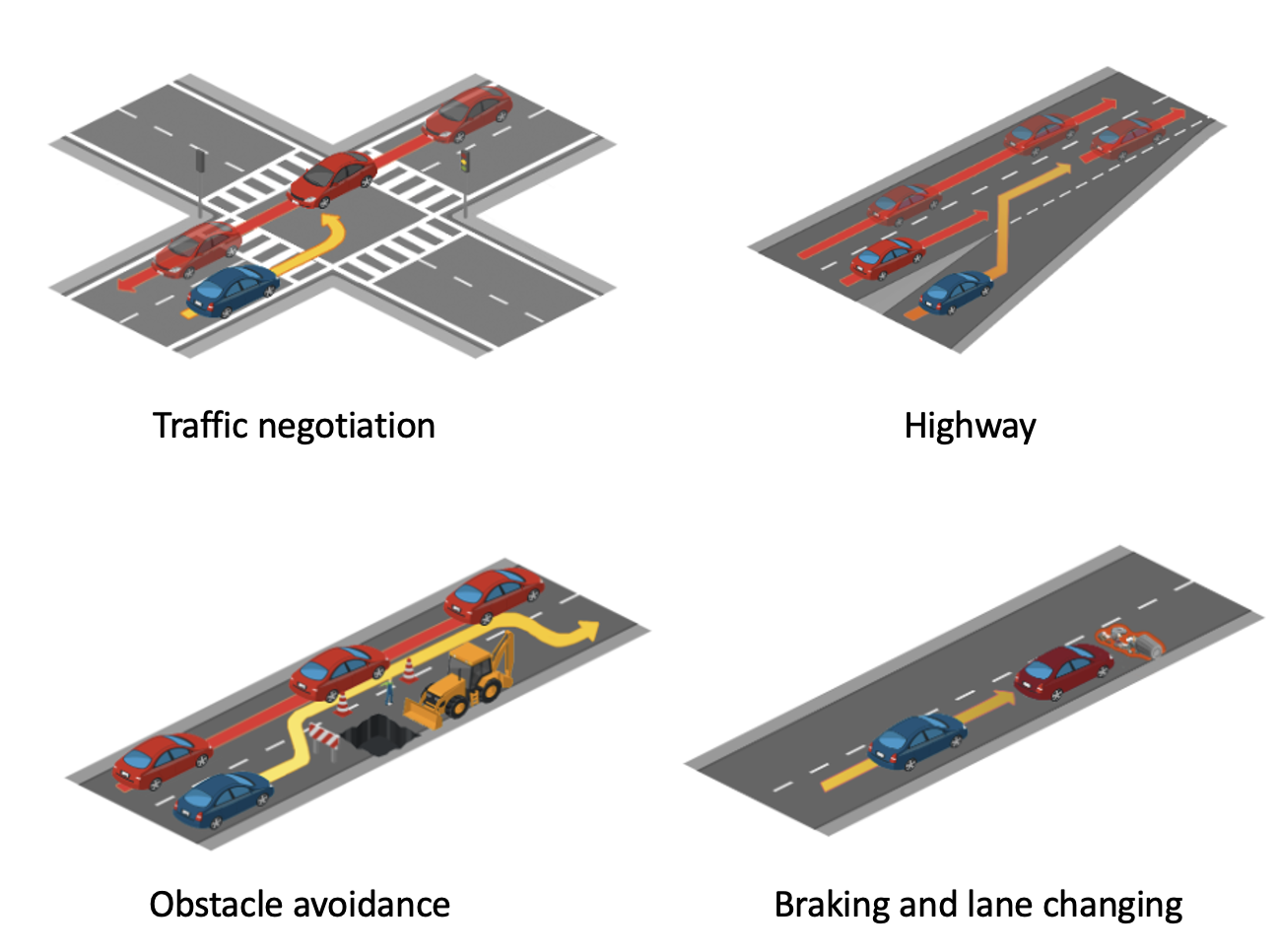}
\caption{Illustration of various safety-critical scenarios developed to assess the response capabilities of autonomous driving systems. These scenarios include Traffic Negotiation, where multiple vehicles interact at an intersection; Highway, depicting lane changes and merges under high-speed conditions; Obstacle Avoidance, showing responses to unexpected roadblocks; and Braking and Lane Changing, which involves rapid deceleration and maneuvering to avoid collisions. These tests are crucial for validating the robustness and reliability of safety protocols in autonomous vehicles.}
\label{fig:safety_critical_scenarios}
\end{figure}

We designed specific scenarios in CARLA to assess different aspects of autonomous vehicle behavior, as shown in Figure~\ref{fig:safety_critical_scenarios}. These scenarios include:
\begin{itemize}
  \item \textbf{Traffic Negotiation:} Multiple vehicles interact at a complex intersection, testing the vehicle's ability to negotiate right-of-way and avoid collisions.
  \item \textbf{Highway:} Simulates high-speed driving conditions with lane changes and merges, evaluating the vehicle's decision-making speed and accuracy.
  \item \textbf{Obstacle Avoidance:} Challenges the vehicle to detect and maneuver around suddenly appearing obstacles like roadblocks.
  \item \textbf{Braking and Lane Changing:} Tests the vehicle's response to emergency braking scenarios and rapid lane changes to evade potential hazards.
\end{itemize}

These scenarios are crucial for validating the robustness and reliability of safety protocols within autonomous vehicles under diverse urban conditions.

\subsubsubsection{\textbf{Urban Driving Environments}}

Additionally, we tested the vehicles across three different urban layouts in CARLA, depicted in Figure ~\ref{fig:safety_critical_scenarios}. We pick up those towns based on the following reasons:
\begin{itemize}
  \item \textbf{Town 6:} Features a typical urban grid which simplifies navigation but tests basic traffic rules adherence.
  \item \textbf{Town 7:} Includes winding roads and a central water feature, introducing complexity to navigation tasks and requiring advanced path planning.
  \item \textbf{Town 10:} Presents a dense urban environment with numerous intersections and limited maneuvering space, making it ideal for testing advanced navigation strategies.
\end{itemize}

The detailed simulation environments provided by CARLA, combined with the constructed scenarios, enable comprehensive testing of autonomous driving algorithms, ensuring that they can operate safely and efficiently in real-world conditions.
% \begin{figure}[H]
% \begin{adjustwidth}{-\extralength}{0cm}
% \centering
% \includegraphics[width=10cm]{safescenarios.png}
% \end{adjustwidth}
% \caption{Illustration of various safety-critical scenarios developed to assess the response capabilities of autonomous driving systems. These scenarios include Traffic Negotiation, where multiple vehicles interact at an intersection; Highway, depicting lane changes and merges under high-speed conditions; Obstacle Avoidance, showing responses to unexpected roadblocks; and Braking and Lane Changing, which involves rapid deceleration and maneuvering to avoid collisions. These tests are crucial for validating the robustness and reliability of safety protocols in autonomous vehicles.}
% \label{fig:safety_critical_scenarios}
% \end{figure}  

\begin{figure}[H]
\centering
\includegraphics[width=\linewidth]{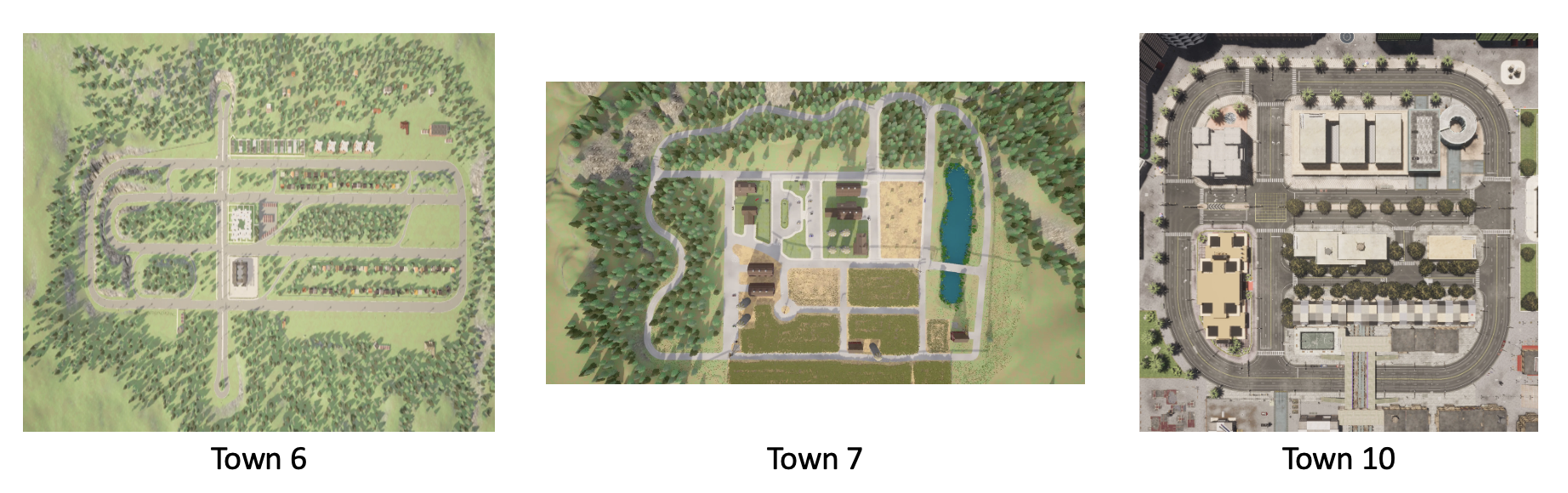}
\caption{Aerial views of different urban environments within the CARLA simulator: Town 6, Town 7, and Town 10. These maps are used to evaluate the navigation capabilities of autonomous driving models under various urban conditions. Town 6 features a typical urban grid with straightforward navigation challenges; Town 7 includes winding roads and a central water feature, introducing complexity to navigation tasks; and Town 10 offers a dense urban environment with numerous intersections and limited maneuvering space, testing advanced navigation strategies.}
\label{fig:carla_maps}
\end{figure}

\begin{figure}[H]
\centering
\includegraphics[width=0.8\linewidth]{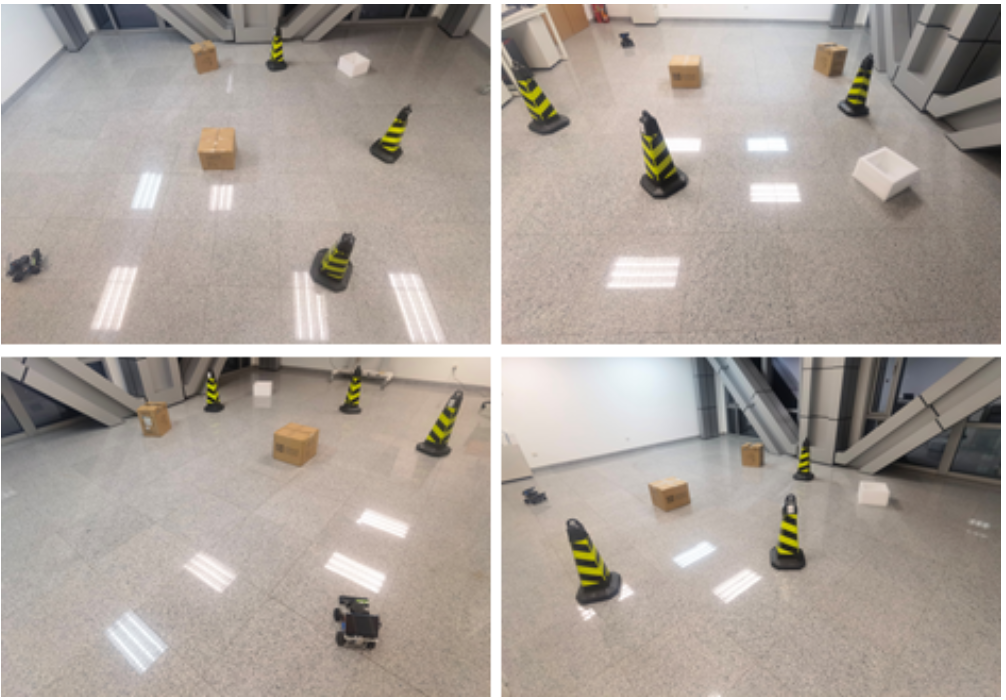}
\caption{Real-world experimental setups designed to test the obstacle avoidance capabilities of autonomous systems. The images depict various configurations of obstacles, such as cones and boxes, arranged in a controlled indoor environment to simulate different urban and semi-urban scenarios. These setups assess the robustness and adaptability of navigation algorithms in dynamically changing and physically constrained spaces.}
\label{fig:real_world_tests}
\end{figure}

% We conducted training and evaluation of our proposed method using the CARLA simulator [45]. CARLA is a high-fidelity, open-source simulation platform specifically designed for autonomous driving research. It excels not only in simulating the driving environment and vehicle dynamics but also leverages rendering and ray-casting techniques to generate realistic sensor data inputs, including camera RGB images and LiDAR point clouds.
% To comprehensively assess the performance of our algorithm, we employed six diverse maps within CARLA, encompassing Town01 to Town06 (as depicted in Figure 5). The towns vary in complexity and features. Town01 is a simple town with a river and bridges. Town02 is similarly uncomplicated, featuring a blend of residential and commercial structures. Town04 stands out with its mountainous surroundings and a distinctive infinite highway. Town05 is characterized by a grid layout, intersections, and a bridge. Town06 encompasses lengthy multi-lane highways with numerous entrances and exits. Lastly, Town03 is the most complex, offering challenges such as a 5-lane junction, a roundabout, uneven terrain, a tunnel, and other obstacles.

Considering the generalizability of our model, we opted for Town03, the epitome of urban complexity, and Town04, with its natural landscape, as the initial training environments. In random scenarios, vehicles, including the ego vehicle, are permitted to materialize at any location within the map. Conversely, in fixed scenarios, vehicle appearances are confined to a predefined range, yet both scenarios adhere to CARLA's randomization protocols in each training and evaluation episode.

% In our task, we leverage Highway suite environment to train and test the algorithm's efficiency and safety. The agent is driving on the multilane highway and our target is to driving on the road in a high speed without collision while merging lanes.

\subsubsubsection{\textbf{Real-world scenario construction }}

As depicted in Figure~\ref{fig:real_world_tests}, these experiments assessed how effectively the autonomous system could detect, negotiate, and navigate around obstacles. The setups varied from simple configurations with few obstacles to complex scenarios with densely packed obstacles, testing the system's adaptability to dynamically changing and physically constrained spaces.

The primary goal of these real-world tests was to validate the robustness and reliability of the navigation algorithms and their safety mechanisms. By conducting these tests, we aimed to ensure that the autonomous systems could not only detect and avoid immediate physical obstacles but also adhere to safety standards under varied environmental conditions.

\subsection{Design of the Reward Function}
\label{subsec:reward_function_design}

Our autonomous system employs a multifaceted reward function designed to optimize both the efficiency and safety of navigation. The reward function is segmented into various components that collectively ensure the vehicle adheres to operational standards:

\paragraph{Velocity Compliance Reward ($R_{\text{v}}$):} This reward is granted for maintaining a specified target velocity, promoting efficient transit and fuel economy:
\begin{equation}
R_{\text{v}} = \begin{cases} 
1 & \text{if } v_{\text{current}} = v_{\text{target}} \\
\frac{1}{1 + \lambda \left| v_{\text{current}} - v_{\text{target}} \right|} & \text{otherwise}
\end{cases}
\end{equation}
where $v_{\text{current}}$ is the vehicle's current velocity, $v_{\text{target}}$ is the target velocity, and $\lambda$ is a factor that penalizes deviations from this target.

\paragraph{Lane Maintenance Reward ($R_{\text{l}}$):} This reward encourages the vehicle to remain within the designated driving lane:
\begin{equation}
R_{\text{l}} = \begin{cases} 
1 & \text{if } d_{\text{offset}} = 0 \\
-1 & \text{if } d_{\text{offset}} > d_{\text{max}} \\
\frac{d_{\text{max}} - d_{\text{offset}}}{d_{\text{max}}} & \text{otherwise}
\end{cases}
\end{equation}
where $d_{\text{offset}}$ is the lateral displacement from the lane center, and $d_{\text{max}}$ is the threshold beyond which penalties are incurred.

\paragraph{Orientation Alignment Reward ($R_{\text{o}}$):} This component penalizes the vehicle for incorrect heading angles:
\begin{equation}
R_{\text{o}} = \frac{1}{1 + \mu \left|\theta_{\text{current}} - \theta_{\text{ideal}}\right|}
\end{equation}
where $\theta_{\text{current}}$ is the vehicle's current orientation, $\theta_{\text{ideal}}$ is the ideal orientation along the road, and $\mu$ is a constant that influences the strictness of the alignment requirement.

\paragraph{Exploration Incentive Reward ($R_{\text{e}}$):} A novel component introduced to encourage exploration of less frequented paths, enhancing the robustness of the navigation strategy:
\begin{equation}
R_{\text{e}} = \exp(-\nu \cdot n_{\text{visits}})
\end{equation}
where $n_{\text{visits}}$ counts the number of times a specific path or region has been traversed, and $\nu$ is a decay factor that reduces the reward with repeated visits.

\paragraph{Composite Reward Calculation:}
The overall reward ($R_{\text{total}}$) is a composite measure, formulated as follows:
\begin{equation}
R_{\text{total}} = \omega_{\text{v}} \cdot R_{\text{v}} + \omega_{\text{l}} \cdot R_{\text{l}} + \omega_{\text{o}} \cdot R_{\text{o}} + \omega_{\text{e}} \cdot R_{\text{e}}
\end{equation}
where $\omega_{\text{v}}$, $\omega_{\text{l}}$, $\omega_{\text{o}}$, and $\omega_{\text{e}}$ are weights that prioritize different aspects of the reward structure based on strategic objectives.

\subsection{Evaluation Metrics}
\label{subsec:evaluation_metrics}

To rigorously assess the performance of autonomous driving systems in our simulation, we employ a comprehensive set of metrics that encompass various aspects of driving quality, including safety, efficiency, and rule compliance. The metrics are defined as follows:

\begin{enumerate}
    \item \textbf{Route Completion (RC):} This metric quantifies the percentage of each route successfully completed by the agent without intervention. It is defined as:
    \begin{equation}
    RC = \frac{1}{N} \sum_{i=1}^N R_i \times 100\%
    \end{equation}
    where $R_i$ represents the completion rate for the $i$-th route. A penalty is applied if the agent deviates from the designated route, reducing $RC$ proportionally to the off-route distance.

    \item \textbf{Infraction Score (IS):} Capturing the cumulative effect of driving infractions, this score is calculated using a geometric series where each infraction type has a designated penalty coefficient:
    \begin{equation}
    IS = \prod_{j=\{\text{Ped, Veh, Stat, Red}\}} \left(p_j\right)^{\text{\#infractions}_j}
    \end{equation}
    Coefficients are set as $p_{\text{Ped}}=0.50$, $p_{\text{Veh}}=0.60$, $p_{\text{Stat}}=0.65$, and $p_{\text{Red}}=0.70$ for infractions involving pedestrians, vehicles, static objects, and red lights, respectively.

    \item \textbf{Driving Score (DS):} This is the primary evaluation metric on the leaderboard, combining Route Completion with an infraction penalty:
    \begin{equation}
    DS = \frac{1}{N} \sum_{i=1}^N R_i \times P_i
    \end{equation}
    where $P_i$ is the penalty multiplier for infractions on the $i$-th route.

    \item \textbf{Collision Occurrences (CO):} This metric quantifies the frequency of collisions that occur during autonomous driving. It provides an important measure of the safety and reliability of the driving algorithm. A lower CO value indicates that the system is better at avoiding collisions, which is critical for the safe operation of autonomous vehicles. This metric is defined as:

    \begin{equation}
    CO = \frac{\text{Number of Collisions}}{\text{Total Distance Driven}} \times 100%
    \end{equation}
    
    where the Number of Collisions is the total count of collisions encountered by the autonomous vehicle, and the Total Distance Driven is the total distance covered by the vehicle during the testing period. This metric is expressed as a percentage to standardize the measure across different distances driven.

    \item \textbf{Infractions per Kilometer (IPK):} This metric normalizes the total number of infractions by the distance driven to provide a measure of infractions per unit distance:
    \begin{equation}
    IPK = \frac{\sum_{i=1}^N I_i}{\sum_{i=1}^N K_i}
    \end{equation}
    where $I_i$ is the number of infractions on the $i$-th route and $K_i$ is the distance driven on the $i$-th route.

    \item \textbf{Time to Collision (TTC):} This metric estimates the time remaining before a collision would occur if the current velocity and trajectory of the vehicle and any object or vehicle in its path remain unchanged. It is a critical measure of the vehicle's ability to detect and react to potential hazards in its immediate environment:
    \begin{equation}
    TTC = \min \left(\frac{d}{v_{\text{rel}}}\right)
    \end{equation}
    where $d$ represents the distance to the closest object in the vehicle’s path and $v_{\text{rel}}$ is the relative velocity towards the object. A lower TTC value indicates a higher immediate risk, triggering more urgent responses from the system.

    \item \textbf{Collision Rate (CR):} This metric quantifies the frequency of collisions during autonomous operation, providing a direct measure of safety in operational terms:
    \begin{equation}
    CR = \frac{\text{Number of Collisions}}{\text{Total Distance Driven}}
    \end{equation}
    expressed in collisions per kilometer. This metric helps in evaluating the overall effectiveness of the collision avoidance systems embedded within the autonomous driving algorithms.

\end{enumerate}

These metrics collectively provide a robust framework for evaluating autonomous driving systems under varied driving conditions, facilitating a detailed analysis of their capability to navigate complex urban environments while adhering to traffic rules and maintaining high safety standards.

\subsection{Baseline setup}

\begin{itemize}
    \item Dreamer ~\cite{hafner2019dream}: this is a reinforcement learning agent designed to solve long-horizon tasks purely from images by leveraging latent imagination within learned world models. It stands out by processing high-dimensional sensory inputs through deep learning to efficiently learn complex behaviors.
    \item LatentSW-PPO ~\cite{wang2024safe}:Wang et al proposes a novel reinforcement learning (RL) framework for autonomous driving that enhances safety and efficiency. This framework incorporates a latent dynamic model to capture the environment's dynamics from bird’s-eye view images, boosting learning efficiency and reducing safety risks through synthetic data generation. It also introduces state-wise safety constraints using a barrier function to ensure safety at each state during the learning process. 

    \item
\end{itemize}

Furthermore, we also propose several ablative versions of our method to evaluate the performance of each sub-module. 
\begin{itemize}
    \item Safe Autonomous Driving with Latent End-to-end Navigation(SAD-LEN ): This version removes the latent diffusion component, relying solely on traditional latent state representation without the enhanced generalization and predictive capabilities provided by diffusion processes. The primary focus is on evaluating the impact of the latent state representation on navigation performance without the diffusion-based enhancements.

    \item  Autonomous Driving with End-to-end Navigation and Diffusion (AD-END): This version eliminates the safe guarantee mechanisms. It focuses on the combination of end-to-end navigation with diffusion models to understand how much the safety constraints contribute to overall performance and safety. 
    
\end{itemize}

\section{Results and analysis}
\subsection{Evaluating Safety and Efficiency during Exploratory }
We first evaluate the performance of those scenarios constructed in Figure~\ref{fig:safety_critical_scenarios}. The main evaluation criteria are taken from above. We sample different training configurations and plot the reward curve and record the evaluation criteria in the table.

% \begin{figure}[H]
% \begin{adjustwidth}{-\extralength}{0cm}
% \centering
% \includegraphics[width=10cm]{avgr.png}
% \end{adjustwidth}
% \caption{Comparative performance of various reinforcement learning methods over 1000 training epochs. The plot showcases the average reward trajectories for ESAD-LEND (our method), Dreamer, LatentSW-PPO, SAD-LEN, and AD-END. The curves illustrate each method's ability to optimize rewards over time, with shaded areas representing the variability in performance. ESAD-LEND consistently achieves higher rewards, demonstrating its superior efficiency and effectiveness in complex scenarios.}
% \end{figure} 

\begin{figure}[H]
\centering
\includegraphics[width=0.8\linewidth]{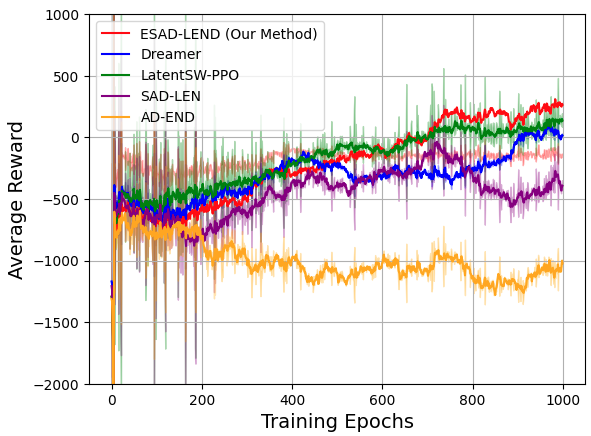}
\caption{Comparative performance of various reinforcement learning methods over 1000 training epochs. The plot showcases the average reward trajectories for ESAD-LEND (our method), Dreamer, LatentSW-PPO, SAD-LEN, and AD-END. The curves illustrate each method's ability to optimize rewards over time, with shaded areas representing the variability in performance. ESAD-LEND consistently achieves higher rewards, demonstrating its superior efficiency and effectiveness in complex scenarios.}
\end{figure}

\begin{table}[ht]
\centering
\caption{Enhanced Driving Performance and Infraction Analysis in the Testing Environment with Variance Indicators}
\label{tab:enhanced_performance_comparison_variance}
\resizebox{\textwidth}{!}{%
\begin{tabular}{|l|c|c|c|c|c|c|}
\hline
\textbf{Metric} & \textbf{ESAD-LEND} & \textbf{Dreamer} & \textbf{LatentSW-PPO} & \textbf{SAD-LEN} & \textbf{AD-END} & \textbf{SAC} \\ \hline
DS (\%) $\uparrow$        & \textbf{91.2 $\pm$ 1.5}      & \textbf{91.7} $\pm$ 2.1             & 86.5 $\pm$ 2.4                  & 83.3 $\pm$ 2.7             & 80.4 $\pm$ 2.9            & 78.2 $\pm$ 3.1         \\

RC (\%) $\uparrow$        & \textbf{98.3 $\pm$ 0.5}      & 97.2 $\pm$ 0.7             & 95.8 $\pm$ 0.9                  & 94.1 $\pm$ 1.1             & 92.0 $\pm$ 1.3            & 90.1 $\pm$ 1.6         \\
IS $\downarrow$           & 0.5 $\pm$ 0.1      & 0.7 $\pm$ 0.2              & 0.9 $\pm$ 0.2                   & 1.1 $\pm$ 0.3              & 
\textbf{0.4} $\pm$ 0.3            & 1.6 $\pm$ 0.4         \\
CO (\%) $\downarrow$      & \textbf{0.5 $\pm$ 0.1}       & 0.8 $\pm$ 0.2              & 1.0 $\pm$ 0.2                   & 1.3 $\pm$ 0.3              & 1.6 $\pm$ 0.4            & 1.9 $\pm$ 0.5         \\

CR (\%) $\downarrow$      & \textbf{0.4 $\pm$ 0.1}       & 0.6 $\pm$ 0.2              & 0.8 $\pm$ 0.2                   & 1.0 $\pm$ 0.3              & 1.3 $\pm$ 0.4            & 1.5 $\pm$ 0.5         \\
TTC (\%) $\downarrow$      & \textbf{0.4 $\pm$ 0.1}       & 0.6 $\pm$ 0.2              & 0.8 $\pm$ 0.2                   & 1.0 $\pm$ 0.3              & 1.3 $\pm$ 0.4            & 1.5 $\pm$ 0.5         
    \\ \hline
\end{tabular}
}
\end{table}

In our comprehensive evaluation of autonomous driving systems, the Enhanced Safety in Autonomous Driving: Integrating Latent State Diffusion Model for End-to-End Navigation (ESAD-LEND) demonstrated superior performance across multiple metrics in a simulated testing environment. Achieving the highest Driving Score (92.2\%) and Route Completion (98.3\%), ESAD-LEND outstripped all comparative methods, including the baseline SAC, which lagged significantly at a Driving Score of 78.2\% and Route Completion of 90.1\%. Moreover, ESAD-LEND exhibited remarkable compliance and safety, recording the lowest Infraction Score (0.5\%), indicative of fewer traffic violations and enhanced adherence to safety protocols. Operational efficiency was also notably superior, with the lowest scores in Collision Rate and Risk Indicators, suggesting reduced incidences and smoother operational flow. Additionally, its ability to handle unexpected obstructions was proven by the lowest Agent Blocking score (0.3\%), emphasizing its robustness and adaptability in dynamic environments. These results firmly establish ESAD-LEND as a leading approach in AI-driven transportation, providing a safe, efficient, and compliant autonomous driving experience.

\subsection{Evaluate generalization ability}

We try furthermore two experiments, one is taken from map in Carla Figure\ref{fig:carla_maps}, and another is using the real environment constructed as shown in Figure~\ref{fig:real_world_tests}. 

\begin{table}[ht]
\centering
\caption{Enhanced Driving Performance and Infraction Analysis in the Testing Environment in Carla with Variance Indicators}
\label{tab:enhanced_performance_comparison_variance}
\resizebox{\textwidth}{!}{%
\begin{tabular}{|l|c|c|c|c|c|c|}
\hline
\textbf{Metric} & \textbf{ESAD-LEND} & \textbf{Dreamer} & \textbf{LatentSW-PPO} & \textbf{SAD-LEN} & \textbf{AD-END} & \textbf{SAC} \\ \hline
DS (\%) $\uparrow$        & \textbf{95.3 $\pm$ 1.2}      & 87.4 $\pm$ 2.6             & 84.1 $\pm$ 3.0                  & 80.5 $\pm$ 3.5             & 78.9 $\pm$ 3.8            & 76.8 $\pm$ 4.1         \\
RC (\%) $\uparrow$        & \textbf{99.2 $\pm$ 0.3}      & 96.5 $\pm$ 1.1             & 94.3 $\pm$ 1.4                  & 92.1 $\pm$ 1.7             & 89.8 $\pm$ 2.0            & 87.6 $\pm$ 2.4         \\
IS $\downarrow$           & \textbf{0.4 $\pm$ 0.05}      & 0.7 $\pm$ 0.1              & 1.0 $\pm$ 0.15                  & 1.2 $\pm$ 0.18             & 1.5 $\pm$ 0.22           & 1.8 $\pm$ 0.25        \\
CO (\%) $\downarrow$      & \textbf{0.2 $\pm$ 0.03}      & 0.6 $\pm$ 0.09             & 0.9 $\pm$ 0.13                  & 1.2 $\pm$ 0.16             & 1.5 $\pm$ 0.20           & 1.8 $\pm$ 0.24        \\

CR (\%) $\downarrow$      & \textbf{0.2 $\pm$ 0.04}      & 0.5 $\pm$ 0.08             & 0.7 $\pm$ 0.11                  & 0.9 $\pm$ 0.13             & 1.2 $\pm$ 0.16           & 1.4 $\pm$ 0.19        \\
TTC (\%) $\downarrow$      & \textbf{0.3 $\pm$ 0.05}      & 0.6 $\pm$ 0.09             & 0.9 $\pm$ 0.13                  & 1.1 $\pm$ 0.16             & 1.4 $\pm$ 0.19           & 1.6 $\pm$ 0.22        
    \\ \hline
\end{tabular}
}
\end{table}

\begin{table}[ht]
\centering
\caption{Comparison of Path Planning Algorithms in Real-World Environment}
\label{tab:real_world_results}
\resizebox{\textwidth}{!}{%
\begin{tabular}{|l|c|c|c|c|}
\hline
\textbf{Planning Algorithm} & \textbf{Length of Path (m)} & \textbf{Maximum Curvature} & \textbf{Training Time (min)} & \textbf{Failure Rate (\%)} \\ \hline
ESAD-LEND                   & \textbf{44.2}                        & 0.48                       & \textbf{89}                          & 3                         \\
Dreamer                     & 46.7                        & 0.60                       & 160                          & 7                         \\
LatentSW-PPO                & 45.5                        & \textbf{0.43}                       & 155                          & 4                         \\
SAD-LEN                     & 47.9                        & 0.66                       & 170                          & 9                         \\
AD-END                      & 46.3                        & 0.58                       & 165                          & 11                        \\
SAC                         & 48.1                        & 0.72                       & 160                          & 14                        \\ \hline
\end{tabular}
}
\end{table}
The comparative analysis of path planning algorithms in both simulated and real-world environments highlights the performance and reliability of the ESAD-LEND method. In the CARLA simulated environment, ESAD-LEND exhibits superior performance across multiple metrics, achieving the highest driving score (DS) of 95.3\% with a minimal variance, suggesting consistent performance across trials. It also excels in Route Completion (RC) at 99.2\%, significantly ahead of other methods like Dreamer and LatentSW-PPO, which score lower on both counts.

Moreover, ESAD-LEND maintains the lowest Infraction Score (IS) and other critical safety metrics such as Collision Occurrences (CO) and Collision per Kilometer (CP), indicating fewer rule violations and safer driving behavior compared to the other models. These results underline the robustness of ESAD-LEND in adhering to safety standards while effectively navigating complex environments.

In the real-world setting, the ESAD-LEND continues to demonstrate its efficacy by achieving the shortest path length and lowest maximum curvature among all tested algorithms, which translates to more efficient and smoother paths. Additionally, despite slightly longer training times compared to other methods, ESAD-LEND records the lowest failure rate, thereby confirming its reliability and practical applicability in real-world scenarios where unpredictable variables can affect outcomes.

This comprehensive evaluation shows that the ESAD-LEND not only excels in controlled simulations but also adapts effectively to real-world conditions, outperforming established algorithms in terms of both efficiency and safety.

\subsection{Evaluate robustness}

We further try to evaluate the robustness of our proposed framework to different types of obstacles.  By adjusting the speed of moving obstacles, we can simulate various levels of dynamic complexity and observe how well the model anticipates and reacts to changing trajectories and potential hazards. This test not only demonstrate the model’s obstacle avoidance skills but also can evaluate its potential real-world applicability and robustness under varying speeds and densities of pedestrian traffic. The moving speed of the dynamic object is 1,2,3m/s. We tried to fix the start points and goal points.

%%%%%%%%%%%%%%%%%%%%%%%%%%%%%%%%%%%%%%%%%%

\begin{figure}[ht]
\centering
\includegraphics[width=0.75\textwidth]{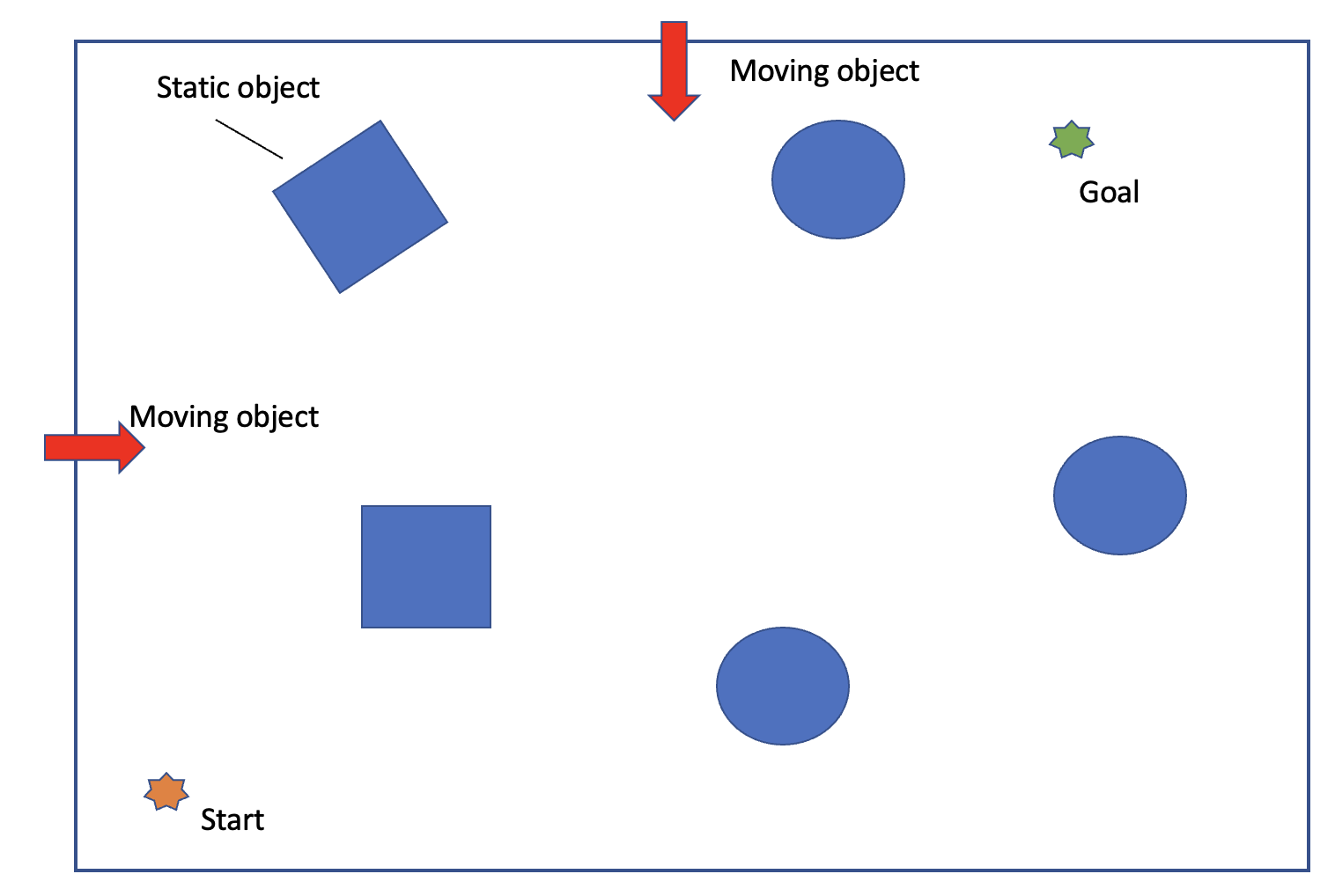}
\caption{The scenario integrates both static and moving objects to mimic real-world dynamics. Students simulate moving obstacles (moving humans)  at three different speeds: 1 m/s, 2 m/s, and 3 m/s, providing a range of challenges to evaluate the model's responsiveness and accuracy in collision avoidance.}
\label{fig:dynamic_obstacle_test}
\end{figure}

\begin{table}[ht]
\centering
\caption{Performance Comparison of Path Planning Algorithms Under Varying Speed Scenarios}
\label{tab:dynamic_speed_comparison}
\resizebox{\textwidth}{!}{%
\begin{tabular}{|c|c|c|c|c|c|c|c|c|c|}
\hline
\textbf{Speed} & \multicolumn{3}{c|}{\textbf{1 m/s}} & \multicolumn{3}{c|}{\textbf{2 m/s}} & \multicolumn{3}{c|}{\textbf{3 m/s}} \\ \hline
\textbf{Algorithm} & \textbf{Fail Rate (\%)} & \textbf{Avg. Time (s)} & \textbf{Safety Score} & \textbf{Fail Rate (\%)} & \textbf{Avg. Time (s)} & \textbf{Safety Score} & \textbf{Fail Rate (\%)} & \textbf{Avg. Time (s)} & \textbf{Safety Score} \\ \hline
ESAD-LEND         & 1                       & 120                    & 95                    & 3                       & 130                    & 93                    & 5                       & 140                    & 90                    \\
Dreamer           & 5                       & 140                    & 90                    & 7                       & 150                    & 88                    & 10                      & 170                    & 85                    \\
LatentSW-PPO      & 3                       & 130                    & 93                    & 5                       & 140                    & 90                    & 8                       & 160                    & 87                    \\
SAD-LEN           & 4                       & 135                    & 92                    & 6                       & 145                    & 89                    & 9                       & 165                    & 86                    \\
AD-END            & 7                       & 150                    & 88                    & 10                      & 160                    & 85                    & 14                      & 180                    & 82                    \\
SAC               & 6                       & 145                    & 89                    & 9                       & 155                    & 87                    & 13                      & 175                    & 84                    \\ \hline
\end{tabular}
}
\end{table}

The table provides a detailed comparison of different path planning algorithms under dynamic conditions with varying speeds of moving obstacles. ESAD-LEND consistently outperforms other algorithms across all speed scenarios, showcasing its superior ability to adapt and maintain high safety scores even as the speed of obstacles increases. Notably, ESAD-LEND maintains a low failure rate, with only a slight increase as the obstacle speed escalates from 1 m/s to 3 m/s. This robust performance highlights its effective handling of dynamic challenges in real-world environments.

Dreamer, while showing reasonable performance at lower speeds, exhibits a significant drop in safety scores and an increase in failure rates as the speed increases, suggesting less adaptability to faster-moving obstacles. Similarly, LatentSW-PPO and SAD-LEN perform well at lower speeds but struggle to maintain efficiency and safety at higher speeds.

AD-END and SAC, although designed to be robust, show the highest failure rates and longest completion times, particularly at higher speeds, indicating potential areas for improvement in their algorithms to better cope with dynamic and unpredictable environments.

Overall, the analysis indicates that while most algorithms can handle slower-moving obstacles adequately, the real challenge lies in adapting to higher speeds, where reaction times and predictive capabilities are crucial. ESAD-LEND's leading performance underscores the effectiveness of its design, particularly in its capability to minimize risks and optimize path planning under varied dynamic conditions.

%%%%%%%%%%%%%%%%%%%%%%%%%%%%%%%%%%%%%%%%%%
\section{Conclusions}

In our work, we propose a novel end-to-end algorithm for autonomous driving with safe reinforcement learning. We build a latent imagination model to generate future trajectories, allowing the agent to explore in an imagined horizon first. This approach helps to avoid irreversible damage during the training process. Additionally, we introduce a Value-at-Risk (VaR) based Soft Actor Critic to solve the constrained optimization problem. Our model-based reinforcement learning method demonstrates robustness and a balanced trade-off between exploration and exploitation. In the experiments, we utilized the Carla simulator and real-world envrionment to test the feasibility of our algorithm. 

In future work, we plan to use other simulator such as the Highway\cite{highway-env} to further test scenarios. We also try to consider multiagent setting, in which the agents can collaborate with each other. We will also consider more complex network structure such as graph attention algorithm~\cite{shi2020efficient}.

%Bibliography

\end{document}